\pdfoutput=1
\documentclass[11pt]{article}
\usepackage{ACL2023}
\usepackage{times}
\usepackage{latexsym}
\usepackage[T1]{fontenc}
\usepackage[utf8]{inputenc}
\usepackage{microtype}
\usepackage{inconsolata}
\usepackage{fancyhdr}
\usepackage{cite}
\usepackage{amsmath,amssymb,amsfonts,amsthm}
\usepackage{algorithmic}
\usepackage{graphicx}
\usepackage{textcomp}
\usepackage{xcolor}
\usepackage{bm}
\usepackage{booktabs}
\usepackage{multirow}
\usepackage{enumitem}
\usepackage{microtype}
\usepackage{graphicx}
\usepackage{sansmath}
\usepackage{pifont}
\usepackage{subfigure}
\usepackage{tcolorbox}
\usepackage{url}

\newcommand{\mc}[1]{{\mathcal #1}}

\newcommand{\cmark}{\ding{51}}%
\newcommand{\xmark}{\ding{55}}%

\newcommand{\eat}[1]{}
\title{A Fused Gromov-Wasserstein Framework for\\ Unsupervised Knowledge Graph Entity Alignment}

\author{Jianheng Tang$^{1,2}$\thanks{~~Work done during an internship at Tencent AI Lab.} , Kangfei Zhao$^{3,4}$, Jia Li$^{1}$\thanks{~~Corresponding author.} \\
  $^{1}$Hong Kong University of Science and Technology (Guangzhou) $^{2}$Hong Kong University of\\ Science and Technology, $^{3}$Tencent AI Lab, $^{4}$Beijing Institute of Technology\\
  \texttt{sqrt3tjh@gmail.com, zkf1105@gmail.com, jialee@ust.hk} \\
}

\begin{document}
\maketitle
\begin{abstract}
Entity alignment is the task of identifying corresponding entities across different knowledge graphs (KGs). Although recent embedding-based entity alignment methods have shown significant advancements, they still struggle to fully utilize KG structural information. In this paper, we introduce FGWEA, an unsupervised entity alignment framework that leverages the Fused Gromov-Wasserstein (FGW) distance, allowing for a comprehensive comparison of entity semantics and KG structures within a joint optimization framework. To address the computational challenges associated with optimizing FGW, we devise a three-stage progressive optimization algorithm. It starts with a basic semantic embedding matching, proceeds to approximate cross-KG structural and relational similarity matching based on iterative updates of high-confidence entity links, and ultimately culminates in a global structural comparison between KGs. We perform extensive experiments on four entity alignment datasets covering 14 distinct KGs across five languages. Without any supervision or hyper-parameter tuning, FGWEA surpasses 21 competitive baselines, including cutting-edge supervised entity alignment methods. Our code is available at \url{https://github.com/squareRoot3/FusedGW-Entity-Alignment}.

\end{abstract}

\section{Introduction}

\begin{figure}[t]
\includegraphics[width=1\linewidth]{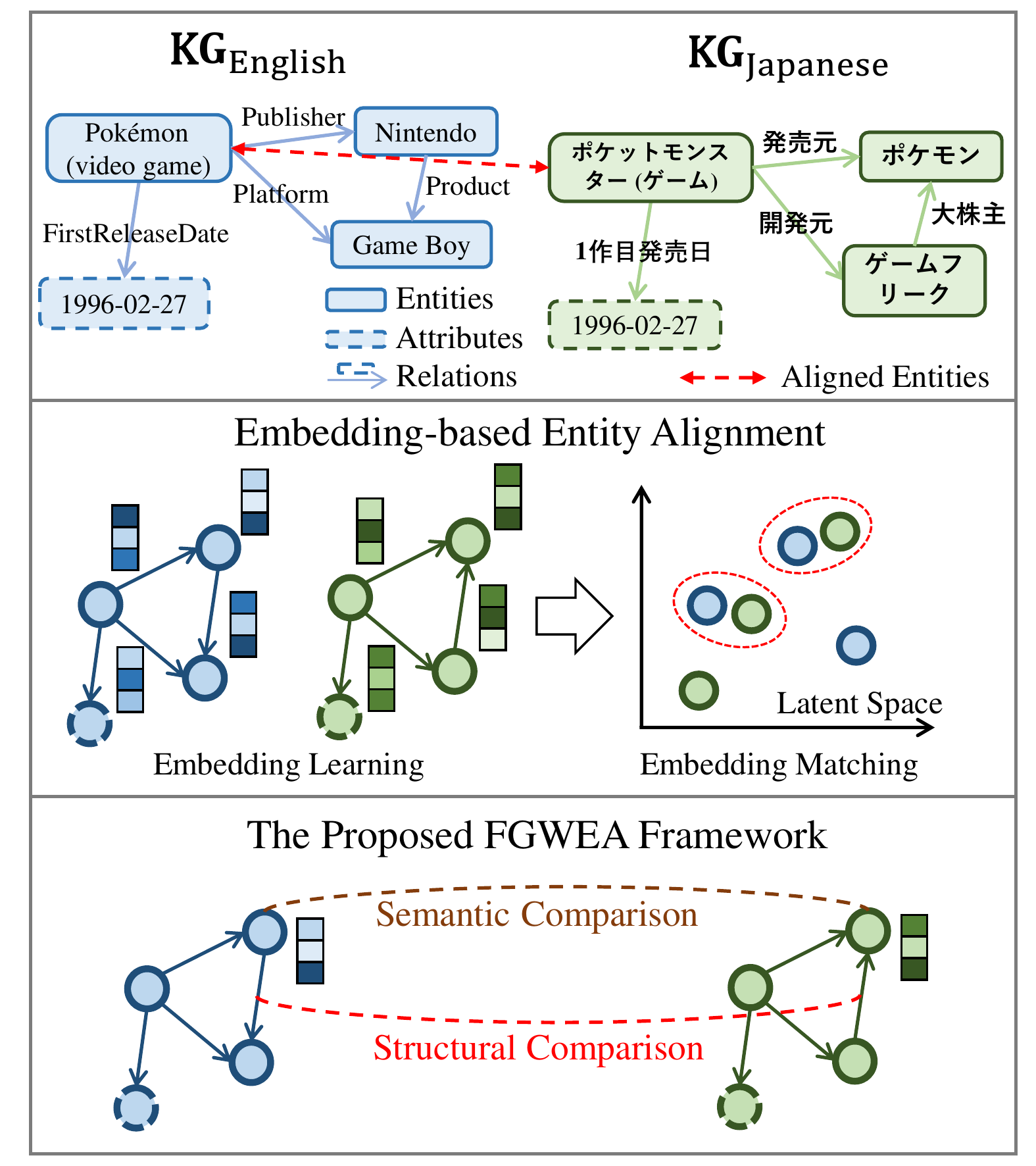}
\caption{Top: A toy example of cross-lingual entity alignment. Middle and bottom: Comparison between embedding-based EA and our proposed FGWEA.}
\label{fig:intro}
\end{figure}

Knowledge Graph (KG) is one of structured data representations that characterizes real-world concepts (also known as entities) with their relationships and attributes. Recent years have witnessed the proliferation of KGs in various areas, ranging from the general ones such as DBpedia \citep{DBpedia} and ConceptNet \citep{conceptnet}, to those in specific domains such as healthcare \citep{healthKG}, education \citep{educationKG}, and e-commerce \citep{commerceKG}. As the information contained in each individual KG is limited and biased, \emph{entity alignment} (EA) is proposed for linking equivalent entities across two KGs from different sources or languages, and integrating them into a new holistic-view KG. EA task has received a lot of attentions in the computational linguistics community, due to its ability to improve the completeness and fairness of KGs, and enhance a wide range of knowledge-driven downstream applications like question-answering \citep{saxena-etal-2020-improving, GeoQA} and dialogue systems \citep{liu2021heterogeneous, xu2019end}. Figure \ref{fig:intro} illustrates a toy example of cross-lingual EA between an English KG and a Japanese KG. The main challenge of this task is to leverage the variety of information in KG, such as entity semantics and relations.

In the deep learning era, embedding-based approaches have become the mainstream for addressing the EA task, which primarily follows the ``embedding-learning-and-matching'' paradigm. As shown in the middle of Figure \ref{fig:intro}, the embedding module encodes entities from two KGs into a shared latent space. The matching module then infers equivalent entities from the embeddings. The basic principle behind embedding-based EA is that equivalent entities in different KGs share similar neighborhood information. Graph neural networks \citep{chang2023knowledge,BWGNN} have been widely adopted as KG encoders, which are usually trained by margin-based losses that encourage equivalent entities to have similar embeddings.

However, the design of the matching module has been overlooked in embedding-based EA. Many existing methods use a greedy strategy that matches entity embeddings to their closest counterparts in another KG, which only relies on the embedding module to incorporate structural information. Unfortunately, even the most powerful KG embedding models and graph neural networks fail to  fully preserve structural information. Although some recent methods have attempted to improve the matching module by treating it as a global assignment problem \citep{SEU} or an optimal transport problem \citep{luo-yu-2022-accurate}, they still fall into the scope of embedding alignment and have limitations in utilizing KG structural information.

To overcome the above issue, we propose FGWEA, an unsupervised EA framework based on the Fused Gromov-Wasserstein (FGW) distance \citep{FusedGW}, which fuses entity embedding alignment (via the Wasserstein distance) and KG structure alignment (via the Gromov-Wasserstein distance) into a joint optimization framework.  As shown in Figure \ref{fig:intro}, instead of only comparing entity embeddings as most embedding-based EA methods did in the literature, the proposed FGWEA jointly incorporates both KG semantics and structure information. In fact, FGWEA considers cross-KG structural and relational consistencies in optimization objectives to better exploit structural information, rather than implicitly encoding it into embeddings. Moreover, after shifting the inclusion of structural information to the matching module and relieving the workload of embedding module, FGWEA is more compatible with pre-trained language models, which only acts as a main tool for encoding semantic information. 

As directly optimizing FGW leads to inefficiency and inferior performance, FGWEA executes a three-stage progressive optimization algorithm, which begins with a relatively simple semantic comparison and then moves on to a more challenging structural comparison. We further develop a fast approximation algorithm and an iterative multi-view OT alignment module to efficiently compare the various KG information. 
Experiments on four cross-lingual and cross-source EA datasets demonstrate that FGWEA outperforms 21 existing EA methods, including both supervised and unsupervised state-of-the-art approaches.

\section{Preliminaries}
\subsection{Task Definition}
\noindent \textbf{Knowledge Graph (KG).} Let $\mc E=\{e_i\}_{i=1}^{|\mc E|}$, $\mc R=\{r_i\}_{i=1}^{|\mc R|}$, $\mc A=\{a_i\}_{i=1}^{|\mc A|}$, $\mc L=\{l_i\}_{i=1}^{|\mc L|}$ be the set of entities, relations, attributes and literals, respectively. Following \citet{qi2021unsupervised}, a KG contains a set of relation triples $\mc T_r=\{(e_i,r_j,e_k)\}$ and attribute triples $\mc T_a=\{(e_i,a_j,l_k)\}$, denoted as $\mc G=(\mc E,\mc R,\mc A,\mc L, \mc T_r,\mc T_a)$. Instances of both types of triples are \textlangle\textit{Pokémon, Publisher, Nintendo}\textrangle ~and \textlangle\textit{Pokémon, FirstReleaseDate, 1996-02-27}\textrangle ~in Figure \ref{fig:intro}. While attribute triples are an essential component in KG, some EA datasets simplify them by only considering the relation triples, i.e., $\mc G=(\mc E,\mc R,\mc T_r)$. Besides, we denote the adjacency matrix of $\mc G$ as $A$, where $A_{ij}=1$ if $e_i$ and $e_j$ connected by at least one relation, and $0$ otherwise.

\paragraph{Entity Alignment (EA).} Given two KGs~ $\mc G$ and $\mc G'$, the EA task is to discover the set of equivalent entity pairs between $\mc G$ and $\mc G'$, denoted as $\mc M=\{(e,e')|e\equiv e',e\in \mc E, e'\in \mc E'\}$, where $e\equiv e'$ means an equivalence relation between $e$ and $e'$. In the unsupervised setting, the EA model predicts $\mc M$ without observing any pre-aligned entities.

\subsection{Optimal Transport (OT)}

The core concept of OT is to find a transportation plan (i.e., the coupling matrix) between two distributions that minimize the overall transportation cost. Let $|\mc E|=m$ and $|\mc E'|=n$; we denote $\mu$ and $\nu$ as two discrete distributions on $\mc E$ and $\mc E'$, respectively. For simplicity, we assume that $\mu$ and $\nu$ follow the uniform distribution. That is, $\mu = \frac{1}{m} \sum_{i=1}^m \delta_{e_i}$ and $\nu = \frac{1}{n}\sum_{j=1}^n \delta_{e'_j}$, where $\delta_{e_i}$ and $\delta_{e'_j}$ are the Dirac measure in $e_i$ and $e'_j$, respectively. We use $\Pi(\mu, \nu)$ to denote the set of all the joint distributions with marginals $\mu$ and $\nu$:
\begin{equation}\label{otdef}
\Pi(\mu,\nu)=\{\pi\ge0: \pi\mathbf{1}_m = \mu, \pi^T\mathbf{1}_n = \nu\},
\end{equation}
where $\pi_{ij}$ signifies the amount of mass transferred from $e_i$ in $\mc G$ to $e'_j$ in $\mc G'$, $\mathbf{1}_m$ denotes an $m$-dimensional all-one vector, and $\pi\mathbf{1}_m$ is the sum of each row in $\pi$. The coupling matrix $\pi$ describes a probabilistic matching of entities between two KGs. A larger value of $\pi_{ij}$ indicates $e_i$ and $e'_j$ are more likely to be aligned. It is worth noting that when $m=n$ and $\mu, \nu$ follow a uniform distribution, \eqref{otdef} corresponds to the ``assignment polytope'', whose vertices correspond to the permutation matrices.
% which covers all valid one-to-one alignment between two groups of n entities.

\paragraph{Wasserstein Distance (WD).} WD is used for directly comparing two distributions, such as two sets of entity embeddings. The Wasserstein distance between $\mu$ and $\nu$ is defined as:
\begin{equation}
\textnormal{WD}(C,\mu,\nu)=\min_{\pi \in \Pi(\mu,\nu)} \sum_{i,j}C_{ij}\pi_{ij},
\end{equation}
where $C_{ij}$ represents the transportation cost between $e_i$ and $e'_j$, e.g., the cosine distance between entity embeddings. We denote the objective in WD as $f_{\textnormal{WD}}(C,\pi)=\sum_{i,j}C_{ij}\pi_{ij}:=\langle C,\pi \rangle$.

\subsection{Gromov-Wasserstein Distance} \label{sec23}

The Gromov-Wasserstein Distance (GWD) \citep{peyre2016gromov} is an extension of the classic OT problem, enabling the alignment of two graphs by solely comparing structures within each graph. Consider $A$ and $A'$ are adjacency matrices of $\mathcal G$ and $\mathcal G'$, GWD is defined as:
\begin{align}\label{GWD}
\textnormal{GWD}(A,A')=&\min_{\pi \in \Pi}\sum_{i,j,k,l}|A_{ij}-A'_{kl}|^2\pi_{ik}\pi_{jl}\nonumber\\
=&\min_{\pi \in \Pi}f_\textnormal{GWD}(A,A',\pi).
\end{align}
In this equation, if $\pi_{ik}$ and $\pi_{jl}$ have large values, it suggests that $(e_i,e'_k)$ and $(e_j,e'_l)$ are likely to be two entity pairs. Consequently, the corresponding intra-KG pairs $(e_i,e_j)$ and $(e'k,e'l)$ should exhibit similar structures, i.e., $|A{ij}-A'{kl}|\to 0$. If two KGs possess identical structures and $\pi$ represents the perfect mapping between them, then $\textnormal{GWD}(A,A')=0$.

\begin{figure*}[t!]
\includegraphics[width=1\textwidth]{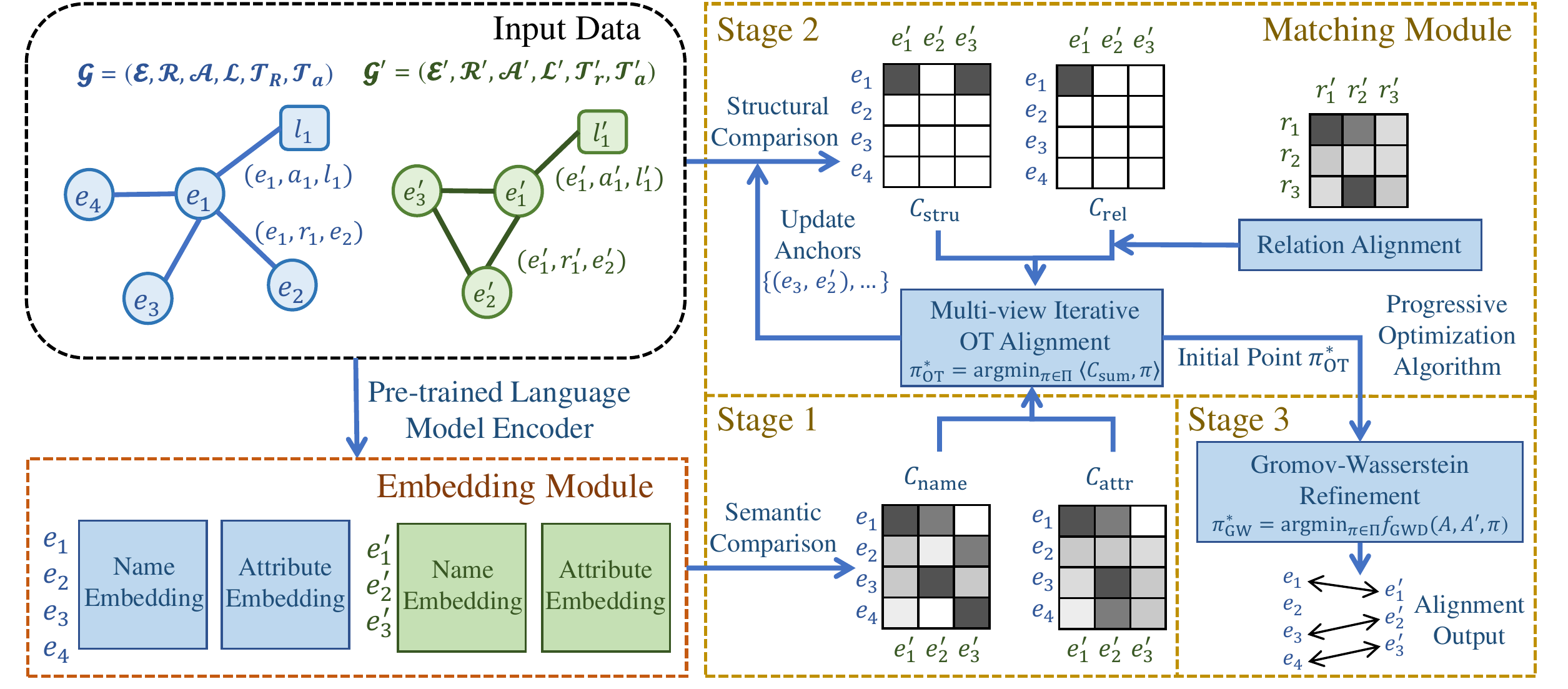} 
\caption{Framework Overview. The embedding module calculates name and attribute embeddings for each entity in KGs. The matching module consists of three stages: semantic comparison (Section \ref{sec31}), multi-view iterative OT alignment (Section \ref{sec32}), and Gromov-Wasserstein refinement (Section \ref{sec33}).}
\label{fig:model}
\end{figure*}

\paragraph{Fused Gromov-Wasserstein Distance (FGW).} Neither WD nor GWD is able to depict the full landscape of KGs. Therefore, FGW \citep{FusedGW} is introduced, whose objective is a linear combination of $f_\textnormal{WD}$ and $f_\textnormal{GWD}$:
\begin{equation}
f_{\textnormal{FGW}}=\alpha f_\textnormal{WD}(C,\pi)+(1-\alpha)f_\textnormal{GWD}(A,A',\pi),
\end{equation}
where $\alpha \in [0,1]$ is a trade-off parameter.

However, several challenges emerge when applying FGW to the EA task. First, GWD assumes that both $A$ and $A'$ are homogeneous graphs, whereas KGs are heterogeneous graphs containing relational information. Second, KG entities possess various forms of side information, such as names and attributes, complicating the accurate measurement of entity similarity and the computation of the cost matrix $C$ in WD. Third, although \citet{FusedGW} invokes the Frank-Wolfe method for optimizing FGW, its effectiveness has only been confirmed on small graphs with hundreds of nodes. We observe that directly applying this method to large-scale sparse KGs results in unstable performance and reduced efficiency. To tackle these issues, we propose a novel EA approach based on FGW in the following section.

\section{The Proposed Method}
We present an unsupervised EA framework, FGWEA, that performs entity matching based on the FGW distance. As shown in Figure \ref{fig:model}, it comprises a semantic embedding module and a three-stage entity matching module. To address the aforementioned challenges, we propose a three-step progressive optimization algorithm. First, FGWEA performs the straightforward semantic embedding matching to obtain high-confidence aligned entity pairs as anchors (Section \ref{sec31}). Building on these anchors, FGWEA employs a fast approximation of GWD to compute cross-KG structural and relational similarities, which are then used for iterative multi-view OT alignment (Section \ref{sec32}). Upon achieving a better initial point for the coupling matrix, FGWEA proceeds to comparing the global structures of KGs by optimizing GWD, the most challenging component in FGW (Section \ref{sec33}).

\subsection{Semantic Embedding and Comparison}\label{sec31}
The embedding module in FGWEA is responsible for encoding entity semantic information, primarily derived from entity names and attributes. Given the remarkable success of pre-trained language models, we employ LaBSE \citep{LaBSE} for embedding multilingual KGs and SimCSE \citep{gao-etal-2021-simcse} for embedding monolingual KGs, both of which are variations of BERT-base \citep{devlin-etal-2019-bert} and are tailored for semantic similarity modeling. It is important to note that our embedding module does not necessitate fine-tuning, and any pre-trained sentence Transformers can be used as a substitute, such as those presented by \citep{reimers-2019-sentence-bert}.

We represent the entity name of $e_i$ as $ne_i$ and concatenated all attribute triples related to $e_i$ into a single string denoted as $ae_i$ (in the form of $a_1l_1a_2l_2\cdots$). The order of the triples depends on the attribute frequency in the KG. Let $\textnormal{enc}(\cdot)$ be the encoder function, we calculate the name similarity-based cost matrix $C_\textnormal{name}$ and attribute similarity-based cost matrix $C_\textnormal{attr}$ between two KGs as follows:
\begin{align}
    C_\textnormal{name} &= 1-\cos(\textnormal{enc}(ne_i),\textnormal{enc}(ne'_j)),\nonumber\\
    C_\textnormal{attr} &= 1-\cos(\textnormal{enc}(ae_i),\textnormal{enc}(ae'_j)).
\end{align}
In the first matching stage, we use the sum of two semantic similarity matrices as the cost in WD and calculate the initial coupling matrix $\pi^0$ by:
\begin{equation}
    \pi^0 = \mathop{\arg\min}_{\pi \in \Pi} \langle C_\textnormal{name} + C_\textnormal{attr}, \pi \rangle.
\end{equation}
Specifically, we use the Sinkhorn algorithm \citep{cuturi2013sinkhorn} to tackle this problem, and collect high confidence entity pairs in $\pi^0$ as \emph{anchors} to facilitate the subsequent matching process. Let $\mc M^0_a$ denote the initial anchor set and $c=1/{\max(m,n)}$ be the maximum potential value of $\pi$. We have $\mc M^0_a=\{(e_i,e'_j)|\pi^0_{ij}>c-\epsilon\}$, where $\epsilon$ is a small threshold satisfying $\epsilon<c/2$ to ensure one-to-one alignment.

\subsection{Approximated GWD for Multi-view Iterative OT Alignment}\label{sec32}

In the second stage, our goal is to incorporate KG structural and relational information into the matching process. Instead of directly optimizing the GWD or FGW objective, we develop an approximate alternative for the sake of efficiency.

\paragraph{Relation-aware GWD} We extebd the structural comparison $|A_{ij}-A_{kl}|^2$ in \eqref{GWD} to the relation comparison:
\begin{equation}\label{relGW}
    \sum_{i,j,k,l}(1-\textnormal{sim}\left(r_{i,j},r'_{k,l})\right)\pi_{ik}\pi_{jl},
\end{equation}
where $r_{i,j}$ represents the relation between $e_i$ and $e_j$. The relation similarity $\textnormal{sim}(r_{i,j},r'_{k,l})=1$ if $A_{ij}=A'_{kl}=1$ and $r_{i,j}\equiv r'_{k,l}$, otherwise 0. As the relation set in different KGs is also unaligned, we align these relations based on relation name similarity, using the same process in Section \ref{sec31}.

\begin{figure}[t]
\includegraphics[width=1\linewidth]{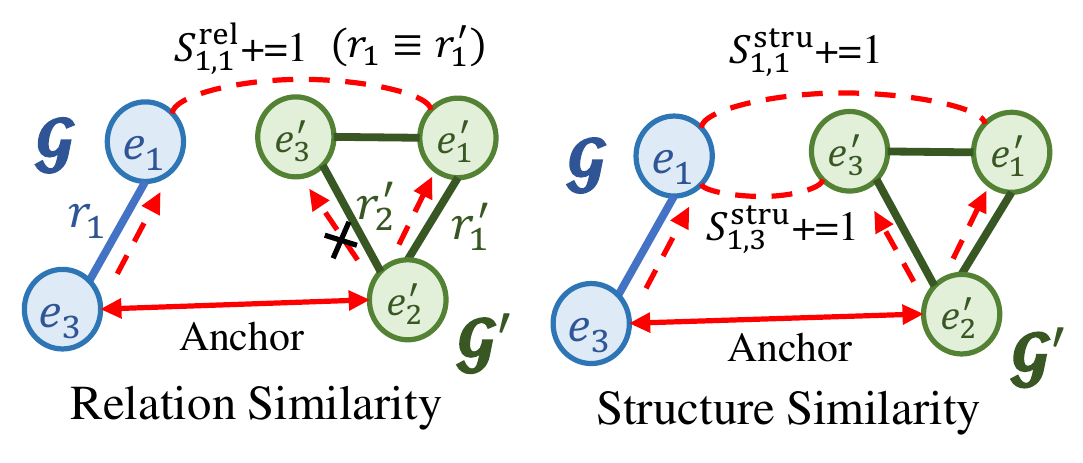} 
% \vspace{-8mm}
\caption{Illustration of how anchor links contribute to $S^\textnormal{stru}$ and $S^\textnormal{rel}$.}
% \vspace{-5mm}
\label{fig:align}
\end{figure}

\paragraph{Approximation} However, optimizing \eqref{relGW} is even more challenging than optimizing GWD. We simplify it by approximating $\pi_{ik}$ in equation \eqref{relGW} with a sparse coupling matrix $\hat \pi$ based on the anchor set $\mc M^0_a$. Specifically, $\hat \pi_{ik}=c$ if $(e_i,e'_k)\in \mc M_a$, and $\hat \pi_{ik}=0$ otherwise. Note that when $\mc M_a$ is closer to the ground truth alignment, the approximation of GWD is more accurate. Afterward, \eqref{relGW} is converted to a WD objective:
\begin{equation}\small
    \sum_{j,l}(1-c\sum_{(e_i,e'_k)\in \mc M_a} \textnormal{sim}(r_{i,j},r'_{k,l}))\pi_{jl}=\langle 1-cS^\textnormal{rel}, \pi\rangle,
\end{equation}
where $S^\textnormal{rel}_{j,l}$ reflects the relation similarity between $e_j$ and $e'_l$. It is calculated by counting the number of anchors $(e_i,e'_k)\in \mc M_a$ in which $e_i$ is a neighbor of $e_j$, $e'_k$ is a neighbor of $e'_l$, and $r_{i,j}\equiv r'_{k,l}$. $S^\textnormal{rel}$ can be efficiently computed by iterating through all anchor pairs and comparing their corresponding neighbor node pairs. Figure \ref{fig:align} illustrates the computation process. If $(e_3,e'_2)$ is an anchor and $e_1,e'_1$ are corresponding neighbors with equivalent relations $r_1 \equiv r'_1$, then $(e_3,e'_2)$ contributes to the relation similarity $S_{1,1}^\textnormal{rel}$. In the right of Figure \ref{fig:align}, we repeat this process to calculate the relation-agnostic structure similarity matrix $S^\textnormal{stru}$, which can be regarded as an approximation of GWD that only compares between anchor entity pairs and other pairs.

\paragraph{Multi-view OT Alignment} To perform a joint comparison of structures and semantics between KGs, we rescale $1-cS^\textnormal{rel}$ and $1-cS^\textnormal{stru}$ to a range of [0,1] and obtain the corresponding cost matrices $C_\textnormal{rel}$ and $C_\textnormal{stru}$. The multi-view OT combines all four cost matrices that represent discrepancies between KGs from different perspectives:
\begin{equation}
    \pi_1^* = \mathop{\arg\min}_{\pi \in \Pi} \langle C_\textnormal{sum}, \pi \rangle,
\end{equation}
where $C_\textnormal{sum}= C_\textnormal{stru}+C_\textnormal{rel}+C_\textnormal{name}+C_\textnormal{attr}$. We derive $\pi_1^*$ and update the anchor set $\mc M^1_a$ with the same process in Section \ref{sec31}. With $\mc M^1_a$, we can adjust $C_\textnormal{stru}$ and $C_\textnormal{rel}$ accordingly, resulting in a new OT problem and a new coupling matrix $\pi^*_2$. We repeat this process for a fixed number of epochs in order to gradually improve the completeness of the anchor set. The final coupling matrix in the second stage is denoted as $\pi_{\textnormal{OT}}^*$.

\subsection{Gromov-Wasserstein Refinement}\label{sec33}
Although the approximated GWD has the advantages mentioned above, the reliance on the anchor set may lead to accumulated error. Therefore, in the final matching stage, we consider the following FGW objective:
\begin{equation}\label{eq:finalFGW}
    f_{\textnormal{FGW}}=\alpha f_{\textnormal{WD}}(C_\textnormal{sum},\pi)+(1-\alpha)f_{\textnormal{GWD}}(A,A',\pi).
\end{equation}
Due to the difficulty in  optimizing $f_{\textnormal{FGW}}$ discussed in \ref{sec23}, we only consider optimizing the second term $f_{\textnormal{GWD}}$ to improve stability. We employ the Bregman Proximal Gradient algorithm, introduced by \citet{xu2019gromov} and shown to have a local linear convergence guarantee by \citet{li2022fast}. For the $k$-th iteration, 
BPG takes the form
\begin{equation}\small
   \pi^{k+1} = \mathop{\arg\min}_{\pi\in \Pi} \{ \nabla_{\pi} f_\textnormal{GWD}(\pi^k)^T \pi + \frac{1}{\beta}\textbf{KL}(\pi||\pi^k) \}, 
\end{equation}
where $\beta$ is the step size and $\textbf{KL}(\cdot||\cdot)$ is the Kullback-Leibler divergence. As such, the $\pi$-update is identical to the entropic OT problem, and we can invoke the Sinkhorn algorithm to tackle it.

Our GW refinement process incorporates two improvements to BPG. First, we use $\pi_{\textnormal{OT}}^*$ as the initial point rather than the uniform distribution, significantly facilitating the optimization process. Second, we employ the relative change of $f_\textnormal{FGW}$ instead of $f_\textnormal{GWD}$ as the optimization stopping criterion, which more accurately reflects the discrepancy between KGs. In the following section, we will test the effectiveness of our proposed FGWEA with the progressive optimization algorithm.

\section{Experiments}

\begin{table}[t!]
% \resizebox{\linewidth}{!}{
\small
\centering
\begin{tabular}{ccccc}  \toprule
Name                        & Lang.     & $|\mc E|$  & $|\mc R|$ & $|\mc T_r|$  \\\midrule
\multirow{2}{*}{DBP15K$_\textnormal{ZH\_EN}$} & ZH & 19,388 & 1,701 & 70,414  \\
                                 & EN & 19,572 & 1,323 & 95,142  \\
\multirow{2}{*}{DBP15K$_\textnormal{JA\_EN}$} & JA & 19,814 & 1,299 & 77,214  \\
                                 & EN & 19,780 & 1,153 & 93,484  \\
\multirow{2}{*}{DBP15K$_\textnormal{FR\_EN}$} & FR & 19,661 & 903   & 105,998 \\
                                 & EN & 19,993 & 1,208 & 115,722 \\\midrule
\multirow{2}{*}{SRPRS$_\textnormal{EN\_FR}$}  & EN & 15,000 & 177   & 33,532  \\
                                 & FR & 15,000 & 221   & 36,508  \\
\multirow{2}{*}{SRPRS$_\textnormal{EN\_DE}$}  & EN & 15,000 & 120   & 37,377  \\
                                 & DE & 15,000 & 222   & 38,363  \\\midrule
\multirow{2}{*}{D-W-15K-V2}      & EN & 15,000 & 167   & 73,983  \\
                                 & EN & 15,000 & 121   & 83,365  \\\midrule
\multirow{2}{*}{Med-BBK-9K}      & ZH & 9,162  & 32    & 158,357 \\
                                 & ZH & 9,162  & 20    & 50,307  \\\bottomrule
\end{tabular}
% }
% \vspace{-2mm}
\caption{Dataset statistics. $|\mc E|$, $|\mc R|$ and $|\mc T_r|$ represent the number of entities, relation types and relation triplets in each KG, respectively.}\label{tab:data}
% \vspace{-5mm}
\end{table}

\begin{table*}[t!]
\resizebox{\textwidth}{!}{
\begin{tabular}{l|cccl|ccc|ccc|ccc}\toprule
    &   \multicolumn{4}{|c}{Configurations}     & \multicolumn{3}{|c|}{DBP15K$_\textnormal{ZH\_EN}$}                  & \multicolumn{3}{c|}{DBP15K$_\textnormal{JA\_EN}$}                  & \multicolumn{3}{c}{DBP15K$_\textnormal{FR\_EN}$} \\
Model      & Name & Attr. & Trans. & Sup. & Hit1           & Hit10          & MRR            & Hit1           & Hit10          & MRR            & Hit1           & Hit10          & MRR            \\\midrule
JAPE       & \xmark   & \checkmark   & \xmark    & 30\%   & 0.412          & 0.745          & 0.490          & 0.363          & 0.685          & 0.476          & 0.324          & 0.667          & 0.430          \\
GCNAlign  & \xmark  & \cmark  & \xmark    & 30\%   & 0.413          & 0.744          & 0.549          & 0.399          & 0.745          & 0.546          & 0.373          & 0.745          & 0.532          \\
\textbf{FGWEA}       & \xmark  & \cmark  & \xmark    & 0\%    & \textbf{0.929} & \textbf{0.978} & \textbf{0.948} & \textbf{0.922} & \textbf{0.974} & \textbf{0.942} & \textbf{0.967} & \textbf{0.994} & \textbf{0.978} \\\midrule
GMatch & \cmark & \xmark   & \xmark    & 30\% & 0.679          & 0.785          & -              & 0.740          & 0.872          & -              & 0.894          & 0.952          & -              \\
SelfKG     & \cmark & \xmark   & \xmark    & 0\%    & 0.745          & 0.866          & -              & 0.816          & 0.913          & -              & 0.957          & 0.992          & -              \\
\textbf{FGWEA}       & \cmark & \xmark   & \xmark    & 0\%    & \textbf{0.926}          & \textbf{0.967}          & \textbf{0.942}          & \textbf{0.954}          & \textbf{0.981}          & \textbf{0.964}          & \textbf{0.996}          & \textbf{0.999}          & \textbf{0.997}          \\\midrule
RDGCN      & \cmark & \xmark   & \cmark   & 30\% & 0.708          & 0.846          & 0.746          & 0.767          & 0.895          & 0.812          & 0.886          & 0.957          & 0.911          \\
DATTI      & \cmark & \xmark   & \cmark   & 0\%    & 0.890          & 0.958          & -              & 0.921          & 0.971          & -              & 0.979          & 0.990          & -              \\
SEU        & \cmark & \xmark   & \cmark   & 0\%    & 0.900          & 0.965          & 0.924          & 0.956          & 0.991          & 0.969          & 0.988          & 0.999          & 0.992          \\
EASY       & \cmark & \xmark   & \cmark   & 0\%    & 0.898          & 0.979          & 0.930           & 0.943          & 0.990          & 0.960           & 0.980          & 0.998          & 0.990           \\
CPL-OT     & \cmark & \xmark   & \cmark   & 0\%    & 0.927          & 0.964          & 0.940           & 0.956          & 0.983          & 0.970           & 0.990          & 0.994          & 0.990           \\
UED      & \cmark & \xmark   & \cmark   & 0\%    & 0.915          & -          &-           & 0.941          & -          & -           & 0.984          & -         & -           \\
LightEA    & \cmark & \xmark   & \cmark   & 0\%    & 0.952          & \textbf{0.984} & 0.964          & 0.981          & \textbf{0.997} & \textbf{0.987} & \textbf{0.995} & 0.998          & \textbf{0.996} \\
\textbf{FGWEA}       & \cmark & \xmark   & \cmark   & 0\%    & \textbf{0.959} & 0.983 & \textbf{0.969} & \textbf{0.982} & 0.995          & \textbf{0.987} & 0.994 & \textbf{0.999} & \textbf{0.996} \\\midrule
AttrGNN    & \cmark & \cmark  & \xmark    & 30\% & 0.796          & 0.929          & 0.845          & 0.783          & 0.921          & 0.834 & 0.919          & 0.978          & 0.910          \\
BERT-INT*  & \cmark & \cmark  & \xmark    & 30\% & 0.968          & 0.990          & 0.977          & 0.964          & 0.991          & 0.975      & 0.992          & 0.998          & 0.995          \\
ICLEA      & \cmark & \cmark  & \xmark    & 0\%    & 0.884          & 0.972          & -              & 0.924          & 0.978          & -              & 0.991          & 0.999          & -              \\
\textbf{FGWEA}       & \cmark & \cmark  & \xmark    & 0\%    & \textbf{0.976} & \textbf{0.994} & \textbf{0.983} & \textbf{0.978} & \textbf{0.992} & \textbf{0.988} & \textbf{0.997} & \textbf{0.999} & \textbf{0.998} \\\midrule
MCLEA*    & \cmark & \cmark  & \cmark   & 30\% & 0.972          & 0.996          & 0.981          & 0.986          & \textbf{0.999} & 0.991          & 0.997          & \textbf{1.000} & 0.998          \\
\textbf{FGWEA}       & \cmark & \cmark  & \cmark   & 0\%    & \textbf{0.987} & \textbf{0.997} & \textbf{0.991} & \textbf{0.991} & 0.998          & \textbf{0.994} & \textbf{0.998} & \textbf{1.000}          & \textbf{0.999} \\\bottomrule
\end{tabular}
}
\caption{Evaluation Results of all compared EA methods on DBP15K under different configurations. Name, Attr., and Trans. represent the usage of entity name, attributes, and translation information, respectively. Sup. indicates the ratio of entity links for supervision. Methods marked with * use additional information not in DBP15K.}
% \vspace{-5mm}
\label{tab:exp1}
\end{table*}

\subsection{Experimental Setup}
\textbf{Datasets.} We evaluate the proposed FGWEA on four frequently used EA datasets, including two multilingual datasets DBP15K \citep{DBP15K_JAPE} and SRPRS \citep{SRPRS_RSN}, and two monolingual multi-source datasets D-W-15K-V2 \citep{OpenEA} and Med-BBK-9K \citep{qi2021unsupervised}. Statistics of these datasets are in Table \ref{tab:data}. For a detailed description, please refer to Appendix \ref{appendix:data}.

\paragraph{Baselines.} A total of 21 EA methods are selected as baselines for performance comparison, spanning from supervised to unsupervised, and conventional to state-of-the-art. Detailed descriptions of most baselines can be found in Section \ref{sec5}. For multilingual EA, we compare with the following methods:  JAPE \citep{DBP15K_JAPE}, GCN-Align \citep{GCNAlign}, GMatch \citep{GMatch}, SelfKG \citep{selfkg}, RDGCN \citep{RDGCN}, DATTI \citep{DATTI}, SEU \citep{SEU}, EASY \citep{EASY}, CPL-OT \citep{CPL-OT}, UED \citep{luo-yu-2022-accurate}, LightEA \citep{LightEA}, AttrGNN \citep{attrGNN}, BERT-INT \citep{BERT-INT}, ICLEA \citep{ICLEA}, and MCLEA \citep{MCLEA}. For monolingual multi-source EA, we compare FGWEA with MultiKE \citep{MultiKE}, BootEA \citep{BootEA}, RSNs \citep{SRPRS_RSN}, LogMap \citep{LogMap}, PARIS \citep{PARIS}, PARSE \citep{qi2021unsupervised}, and StrMatch, a simple matching method using the string edit distance.

\paragraph{Evaluation Metrics.} On DBP15K and SPARS, we use \textbf{HitK} and \textbf{MRR} to evaluate the performance of all EA methods. \textbf{HitK} calculates the percentage of entities in $\mc G$ whose counterparts in $\mc G'$ is in the top-K candidates of model output. \textbf{MRR} is the mean reciprocal rank. On D-W-15K-V2 and Med-BBK-9K, we adopt another evaluation protocol for a comprehensive evaluation suggested by \citet{leone2022critical}. We use the standard classification-based metrics, i.e., precision (\textbf{P}), recall (\textbf{R}), and $\bm F_1$ scores between the set of all predicted entity pairs and that of ground truth entity pairs.

\paragraph{Implementation Details.} Unlike most neural-based EA methods, the proposed FGWEA requires no hyper-parameter tuning and we use the same hyper-parameters across all datasets. We update 6 epochs for multi-view OT alignment and set the threshold $\epsilon$ to 1e-5. In all places where the Sinkhorn algorithm is used, we set the entropic regularization weight $\eta$ to 0.1 and the number of iterations to 10. We set $\alpha$ in the FGW objective \eqref{eq:finalFGW} to be the average graph density of $A$ and $A'$ to maintain a balance between the magnitude of the WD and GWD terms. We set the step size $\beta$ in BPG to 100 and the maximum iteration number to 2000. The only exception is that we encounter numerical errors on the Med-BBK-9K dataset, and thus decrease $\beta$ to 50. Our model is implemented on PyTorch. All experiments are performed on a Linux server with an AMD Ryzen9 5950X CPU and an NVIDIA GeForce RTX 3090 GPU. 

\subsection{Results on Cross-lingual EA Datasets}

DBP15K is the most widely-adopted EA dataset. Unfortunately, the experimental configurations of different baselines on this dataset are highly inconsistent, leading to unfair comparison. 
After a careful study of existing work, we figure out four factors that significantly effect the results: (1) the inclusion of entity names, (2) the utilization of attribute triples, (3) the use of Google translation for non-English entities, and (4) the ratio of entity links for supervision. 

Based on factors (1-3), we categorize baselines into five groups and run FGWEA using the configurations for each group. The experimental settings and results of all compared baselines and FGWEA is in Table \ref{tab:exp1}. As observed, FGWEA achieves the best performance in terms of Hit1 and MRR in all five groups. Specifically, the \emph{unsupervised} FGWEA outperforms two state-of-the-art \emph{supervised} EA approaches BERT-INT and MCLEA. SelfKG and ICLEA are two graph neural network-based methods that use the same pre-trained language model named LaBSE to encode semantic information. However, our approach outperforms them by a significant margin, demonstrating its ability of utilizing KG structures. UED and CPL-OT, which are also based on OT for alignment, do not perform as well as FGWEA, suggesting that the FGW distance we introduced is more suitable for this task.

\begin{table}[t]
\centering
\resizebox{0.9\linewidth}{!}{
\begin{tabular}{l|cc|cc}\toprule
              & \multicolumn{2}{c|}{SRPRS$_\textnormal{EN\_FR}$}                     & \multicolumn{2}{c}{SRPRS$_\textnormal{EN\_DE}$}                     \\
Model        & Hit1            & Hit10          & Hit1            & Hit10          \\\midrule
% RDGCN*        & 0.672           & 0.767          & 0.779           & 0.886          \\
BERT-INT      & 0.971           & 0.975          & 0.986           & 0.988          \\\midrule
CPL-OT        & 0.974           & 0.988          & 0.974           & 0.989          \\
EASY*         & 0.965           & 0.989          & 0.974           & 0.992          \\
SEU*          & 0.982           & 0.995          & 0.983           & 0.996          \\
LightEA*      & 0.986           & 0.994          & 0.988           & 0.995          \\\midrule
\textbf{FGWEA} & \textbf{0.996}  & \textbf{0.999} & \textbf{0.997}  & \textbf{1.000} \\\bottomrule
\end{tabular}
}
% \vspace{-2mm}
\caption{Evaluation Results on the SPARS dataset. Methods marked with * used the translated entity name.}\label{tab:exp2}
% \vspace{-2mm}
\end{table}

Table \ref{tab:exp2} reports the results on the SPARS dataset. BERT-INT uses 30\% entity links for training and other baselines are unsupervised. While most baselines rely on translated entity names to overcome the language barrier, FGWEA achieves the best performance with untranslated entity names. It surpasses LightEA, the current leading method on this dataset, by reducing the error rate from 1.2\% to only 0.3\% on SPRPS$_\textnormal{EN\_DE}$.

\subsection{Results on Cross-source EA Datasets}

\begin{table}[t]
\centering
\small
\resizebox{\linewidth}{!}{
\begin{tabular}{l|ccc|ccc}\toprule
        & \multicolumn{3}{c|}{D-W-15K-V2}                & \multicolumn{3}{c}{MED-BBK-9K}                     \\
Model   & P             & R             & $F_1$            & P             & R             & $F_1$            \\\midrule
MultiKE & 49.5          & 49.5          & 49.5          & 41.0          & 41.0          & 41.0          \\
BootEA  & 82.1          & 82.1          & 82.1          & 30.7          & 30.7          & 30.7          \\
RSNs    & 72.3          & 72.3          & 72.3          & 19.5          & 19.5          & 19.5          \\\midrule
StrMatch  & 60.6          & 41.9          & 49.5          & 54.5          & 49.5          & 51.9          \\
LogMap  & -             & -             & -             & 86.4          & 44.1          & 58.4          \\
PARIS   & 95.0          & 85.0          & 89.7          & 77.9          & 36.7          & 49.9          \\
PRASE   & 94.8          & 90.0          & 92.3          & 83.7          & 61.9          & 71.1          \\\midrule
\textbf{FGWEA}    & \textbf{95.2} & \textbf{90.3} & \textbf{92.7} & \textbf{93.9} & \textbf{73.2} & \textbf{82.3} \\\bottomrule
\end{tabular}
}
% \vspace{-2mm}
\caption{Results on cross-source EA datasets.}\label{tab:exp3}
% \vspace{-5mm}
\end{table}

Cross-source EA poses more challenges than EA within the same knowledge source due to the larger discrepancies in schema and topology of KGs from different sources. For example, in D-W-15K-V2, we find the KG from WikiData uses OIDs as entity names. To facilitate semantic comparison in FGWEA, we replace these OIDs with entity attributes that possess linguistic information.

In Table \ref{tab:exp3}, we compare FGWEA with 7 EA methods that were not included in cross-lingual EA evaluation. The results show that FGWEA consistently outperforms all the baselines on two datasets in terms of precision, recall, and $F_1$ scores. Remarkably, FGWEA outperforms PARSE by 11.9\% in terms of $F_1$, which is the previous best performed method on this dataset. FGWEA also surpasses PARIS, a conventional approach that has shown superior performance to all neural-based EA in a recent study \citep{leone2022critical}.

\subsection{Ablation Study and Model Efficiency}
\begin{table*}[t!]
\resizebox{\textwidth}{!}{
\begin{tabular}{l|cccr|cccr|rrrr|rrrr}\toprule
            & \multicolumn{4}{c|}{DBP15K$_\textnormal{ZH\_EN}$}                        & \multicolumn{4}{c|}{SRPRS$_\textnormal{EN\_FR}$}                        & \multicolumn{4}{c|}{D-W-15K-V2}                                           & \multicolumn{4}{c}{MED-BBK-9K}                     \\
Model       & Hit1           & Hit10          & MRR            & Time & Hit1           & Hit10          & MRR            & Time & P             & R             & $F_1$            & Time           & P             & R             & $F_1$            & Time                 \\\midrule
FGWEA   & \textbf{0.987} & \textbf{0.997} & \textbf{0.991} & 254  & \textbf{0.996} & \textbf{0.999} & \textbf{0.997} & 388  & 95.2          & \textbf{90.3} & \textbf{92.7} & 494                      & \textbf{93.9} & \textbf{73.2} & \textbf{82.3} & 151                  \\
- w/o GW     & 0.975          & 0.992          & 0.981          & 57   & 0.979          & 0.989          & 0.983          & 62   & 95.8          & 84.7          & 89.9          & 49                       & 92.4          & 58.5          & 71.7          & 19                   \\
- w/o $S_{rel}$  & 0.970          & 0.990          & 0.977          & 52   & 0.976          & 0.987          & 0.980          & 49   & \textbf{98.7} & 82.2          & 90.0          & 44                       & 92.6          & 57.8          & 71.1          & 18                   \\
- w/o $S_{stru}$ & 0.951          & 0.979          & 0.962          & 40   & 0.966          & 0.982          & 0.972          & 23   & 97.5          & 56.6          & 71.6          & 25                       & 89.4          & 45.2          & 60.1          & 10                   \\
\midrule
GW-only     & 0.011          & 0.026          & 0.017          & 555  & 0.004          & 0.023          & 0.011          & 504  & 0.3           & 0.1           & 0.1           & 227                      & 0.3           & 0.1           & 0.1           & 470                  \\
Emb-Match   & 0.763          & 0.861          & 0.799          & 1    & 0.915          & 0.959          & 0.931          & 1    & 57.3          & 57.3          & 57.3          & 1                        & 51.1          & 51.1          & 51.1          & 1            \\\bottomrule       
\end{tabular}
}
% \vspace{-2mm}
\caption{Ablation study of FGWEA. The wall-clock time is measured in seconds.}
% \vspace{-4mm}
\label{tab:exp4}
\end{table*}

To validate the effectiveness and efficiency of each component in FGWEA, we compare it with several ablations. First, we remove Gromov-Wasserstein refinement, the third matching stage in FGWEA, and refer to this new version as FGWEA w/o (without) GW. Then, we continue to remove the relational comparison and structural comparison in the second matching stage, and obtain FGWEA w/o $C_\textnormal{rel}$ and $C_\textnormal{stru}$, respectively. GWD-only is a baseline that directly optimizes GWD for alignment without using the progressive optimization algorithm in FGWEA. Emb-Match directly matches entities based on entity semantic embeddings. 

As shown in Table \ref{tab:exp4}, FGWEA performs the best compared with these variants, which validates the effectiveness of the proposed progressive optimization algorithm. Removing GW refinement in FGWEA results in a decrease in performance on all datasets and a significant reduction in computational time. Removing either the relational comparison or the structural comparison also leads to a decline in performance, while the time consumption does not change significantly. Besides, directly optimizing GWD between KG structures is ineffective for the EA task, and aligning entity semantic embeddings alone also has poor performance. This highlights the importance of considering structural and semantic information jointly.

Note that Table \ref{tab:exp4} only calculates the time spent on the matching module. The embedding module takes approximately 5 minutes to run on DBP15K and 3 minutes on other datasets. On average, it takes approximately 10 minutes to run FGWEA on these datasets, which is relatively efficient compared to most embedding-based EA methods. 

\subsection{Visualization of the FGW Objective}\label{sec45}

\begin{figure}[t]
\centering
\includegraphics[width=0.9\linewidth]{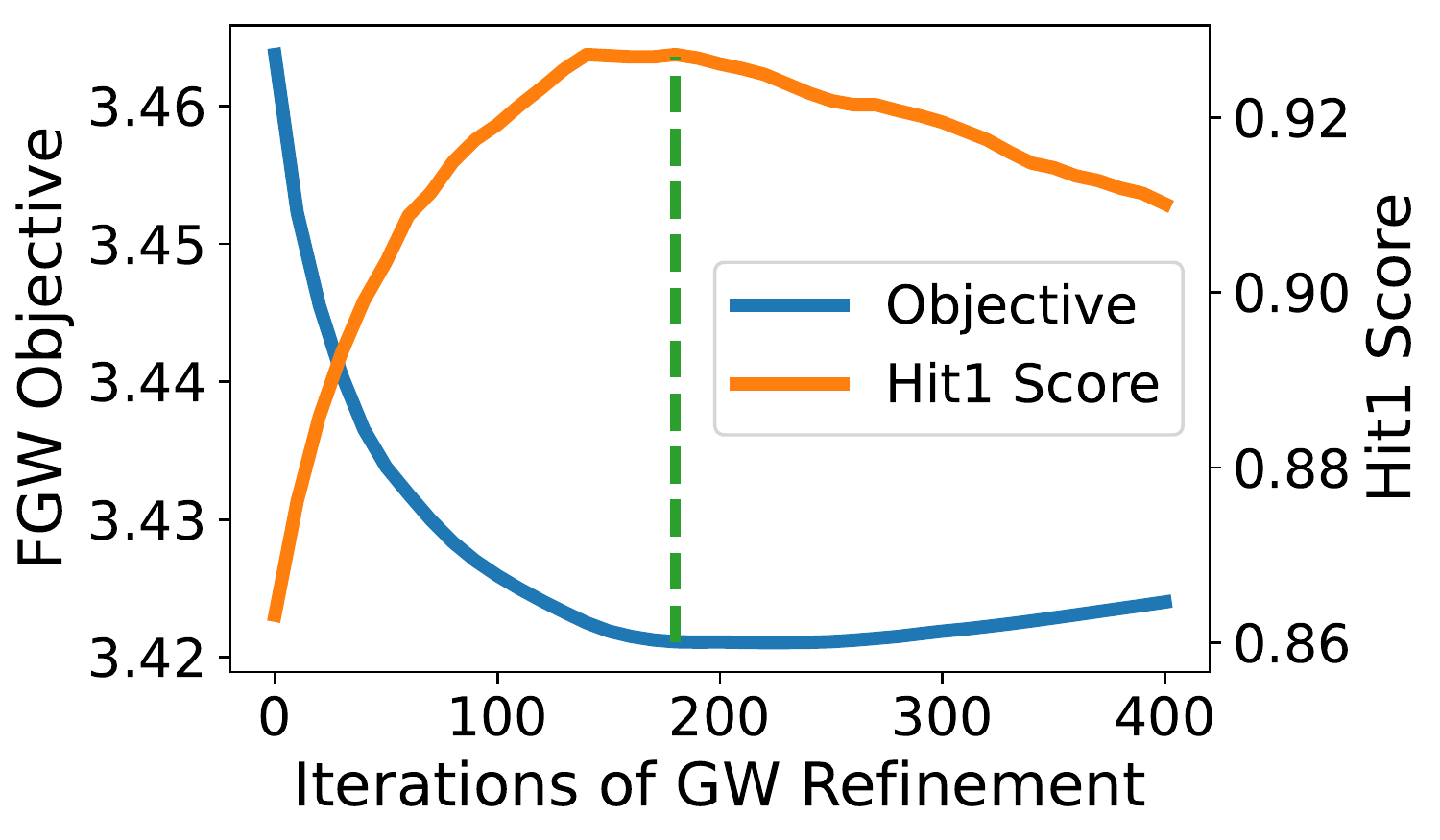} 
\caption{Visualization of the relationship between the objective function and alignment performance (Hit1) of FGWEA in the GW refinement process.}
\label{fig:exp}
\end{figure}

In figure \ref{fig:exp}, we visualize the objective function in \eqref{eq:finalFGW} and the corresponding Hit1 score for 400 epochs in GW refinement on DBP15K$_\textnormal{ZH\_EN}$ without translation and attributes. We find a strong correlation between two curves---the iteration corresponding to the minimum FGW objective value is approximately that to the maximum Hit1 score. This suggests that the FGW objective can be utilized as an unsupervised metric to estimate the alignment performance and to help determine when to stop optimization in GW refinement. In this case, WD increases monotonically, and GWD decreases monotonically in all steps, neither of which are able to indicate model performance. More examples can be found in Appendix \ref{appendix:case}.

\section{Related Work}\label{sec5}
\subsection{Unsupervised Entity Alignment}
\eat{Unsupervised EA methods align entities across two KGs without the need for labeled entity pairs, making them particularly useful in real-world scenarios where labels are expensive or unavailable. We categorize the existing methods into three groups.}
We categorize the existing unsupervised entity alignment methods into three groups:

\noindent(1) \textbf{Traditional heuristic EA systems}. LogMap \citep{LogMap} and PARIS \citep{PARIS} are two well-known traditional EA systems that iteratively discover entity links by logical inference, lexical matching, and probabilistic reasoning. PARSE \citep{qi2021unsupervised} is an enhanced version of PARIS which combines probabilistic reasoning and semantic embedding. 

\noindent(2) \textbf{Self-supervised neural EA methods}. SelfKG \citep{selfkg} uses the graph neural network to aggregate entity embeddings of one-hot neighbors, and proposes a similarity metric between the entities of two KGs for contrastive learning. ICLEA \citep{ICLEA} conducts bidirectional contrastive learning via building pseudo-aligned entity pairs as pivots for cross-KG interaction.

% EASY \citep{EASY} includes an entity name alignment module and an iterative matching module that fuses the name and structural information to correct the misaligned entities. 
\noindent(3) \textbf{Optimization-based non-neural EA methods}. SEU \citep{SEU} transforms the EA problem into assignment problem. LightEA \citep{LightEA} is a non-neural framework which reinvents the label propagation algorithm to effectively run on KGs. Our proposed FGWEA also belongs to this group.

\subsection{Optimal Transport for Entity Alignment}
There have been a few approaches that use OT to improve the EA performance. OTEA \citep{OTEA} is a supervised method that adopts the basic TransE \citep{TransE} for KG embedding and proposes the group-level loss for embedding training based on OT theory. \citep{luo-yu-2022-accurate} develops a modified OT problem for global EA and dangling entity detection. CPL-OT \citep{CPL-OT} employs a graph convolutional network to learn entity embeddings, which are then utilized to determine the transportation cost in OT and resolve alignment conflicts. SLOTAlign \citep{SLOTAlign} is an unsupervised graph alignment framework that jointly performs structure learning and optimal transport alignment. Compared with these methods, FGWEA takes the first step towards introducing the Fused Gromov-Wasserstein distance to EA, which enables better utilization of the structural information in KGs.

\section{Conclusion}
In this paper, we propose an unsupervised entity alignment framework named FGWEA. Instead of following the ``embedding-learning-and-matching'' paradigm, we invoke the Fused Gromov-Wasserstein distance to realize a more explicit and comprehensive comparison of structural and semantic information between knowledge graphs. To realize the benefits of FGWEA, we present a three-stage progressive optimization algorithm to address the challenge of optimizing the FGW objective. Experimental results show that FGWEA outperforms both supervised and unsupervised state-of-the-art entity alignment methods.

\section*{Acknowledgement}
This research was supported by NSFC Grant No. 62206067, Tencent AI Lab Rhino-Bird Focused Research Program RBFR2022008 and Guangzhou-HKUST(GZ) Joint Funding Scheme 2023A03J0673.

\section*{Limitations}
Although the proposed FGWEA framework demonstrated the superior performance on multiple public EA datasets, there are still some limitations that require further research.

\paragraph{Scalability.} In this paper, we have successfully extended FGW to KGs with tens of thousands of entities, which is the common size of domain-specific KGs. However, real-world general-domain KGs can be much larger and contain millions of entities. The most time-consuming step in FGWEA, the Gromov-Wasserstein refinement, has quadratic time complexity $O(|\mc E||\mc T'_r|+|\mc E'||\mc T_r|)$ and thus cannot be directly applied to million-scale KGs. There are three ways to further scale up FGWEA. First, we can remove the most time consuming step, GW refinement, while FGWEA still has competitive performance in Table \ref{tab:exp4}. Second, we can use recent divide-and-conquer methods \citep{xin2022large, zeng2022entity,li2021mask} to divide large scale KGs into smaller subgraph pairs, and then apply alignment methods for each subgraph pair. Third, the coupling matrix $\pi$ can be restricted to a sparse matrix which only considers top-k candidates for each entity, and the computation can be accelerated by mask OT \citep{gasteiger2021scalable} or sparse Sinkhorn iteration \citep{LightEA}.

\paragraph{Dealing with dangling cases.} FGWEA supposes all entities have equal probabilities to be matched in the beginning by using the uniform distribution. Therefore, it has limited ability to handle dangling entities whose counterparts are unavailable in the other KG \citep{sun-etal-2021-knowing}. To avoid this limitation, we can invoke unbalanced OT \citep{UOT} or unbalanced GWD \citep{UGW}, which relax the assumption of equal probabilities for all entities.

\bibliography{ref}
\bibliographystyle{acl_natbib}
\clearpage
\appendix
\section{Dataset Description}\label{appendix:data} All the datasets used in our evaluation are publicly available on the Internet. 

\noindent\textbf{DBP15K}\footnote{\url{https://github.com/nju-websoft/JAPE}} consists of three subsets of cross-lingual KG pairs extracted from DBpedia: DBP15K$_\textnormal{ZH\_EN}$ (Chinese to English), DBP15K$_\textnormal{JA\_EN}$ (Japanese to English), and DBP15K$_\textnormal{FR\_EN}$ (French to English). Each KG pair contains 15,000 pre-aligned entity links. 

\noindent\textbf{SRPRS}\footnote{\url{https://github.com/nju-websoft/RSN}} is a sparse dataset that includes two cross-lingual KG pairs extracted from DBpedia: SRPRS$_\textnormal{EN\_FR}$ (English to French), and SRPRS$_\textnormal{EN\_DE}$ (English to German). Each subset of SRPRS also contains 15,000 entity links, but with fewer relation triples and no attribute triples. 

\noindent\textbf{D-W-15K-V2}\footnote{\url{https://github.com/nju-websoft/OpenEA}} consists of two English KGs extracted from DBpedia and WikiData, respectively, and there are 15,000 pre-aligned entity links. 

\noindent\textbf{MED-BBK-9K}\footnote{\url{https://github.com/ZihengZZH/industry-eval-EA}} is an industry dataset containing two Chinese medical KGs with 9,162 entity links, one is an authoritative human annotated KG and the other is extracted from a Chinese online encyclopedia called Baidu Baike. D-W-15K-V2 is licensed under the GNU General Public License v3.0, while other datasets are licensed under the MIT License.

\begin{figure}[t!]
\includegraphics[width=1\linewidth]{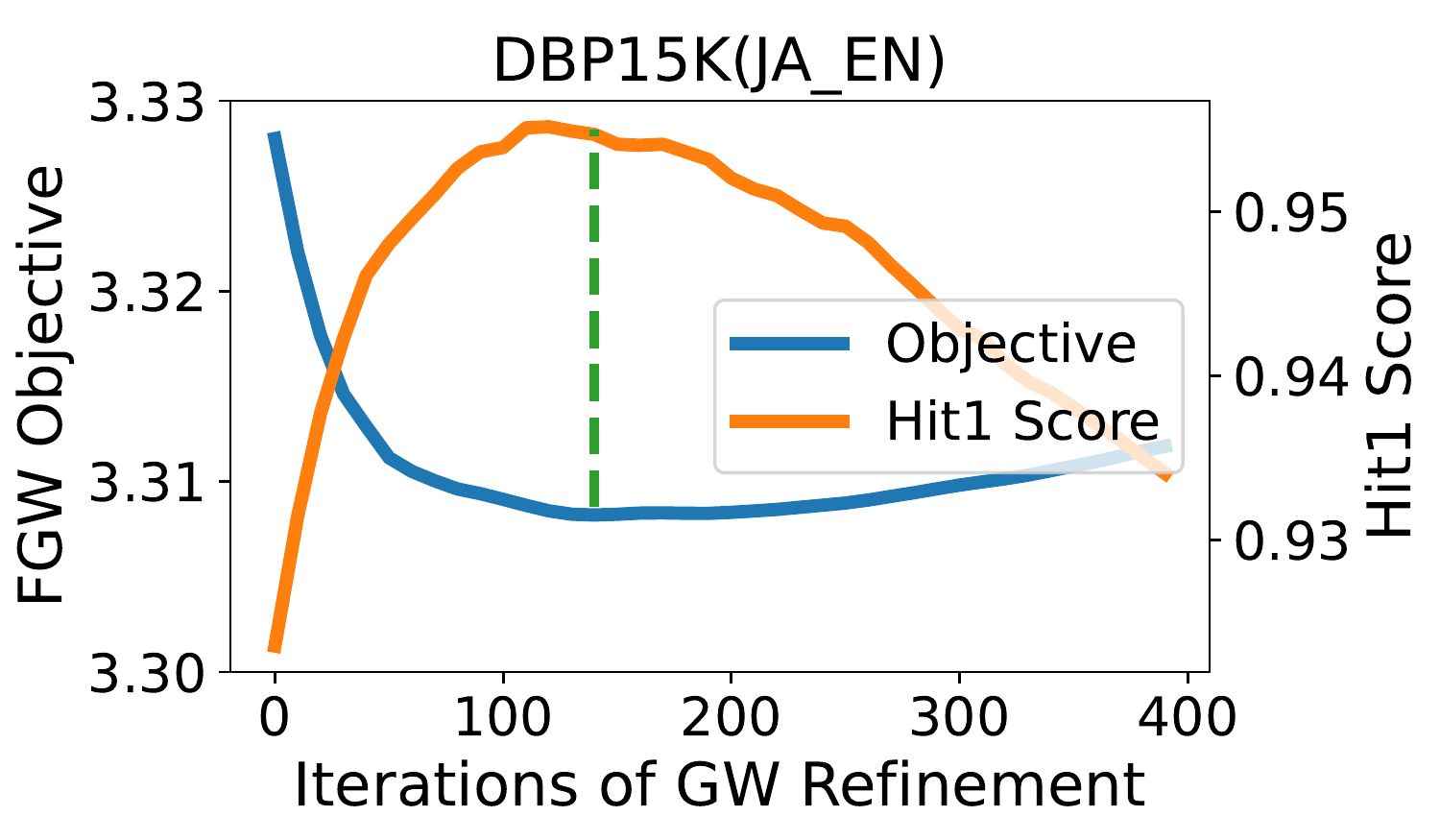} 
\caption{Visualization of the relationship between the objective function and alignment performance on DBP15K$_\textnormal{EN\_JA}$.}
\label{fig:exp2}
\end{figure}

\begin{figure}[t!]
\includegraphics[width=1\linewidth]{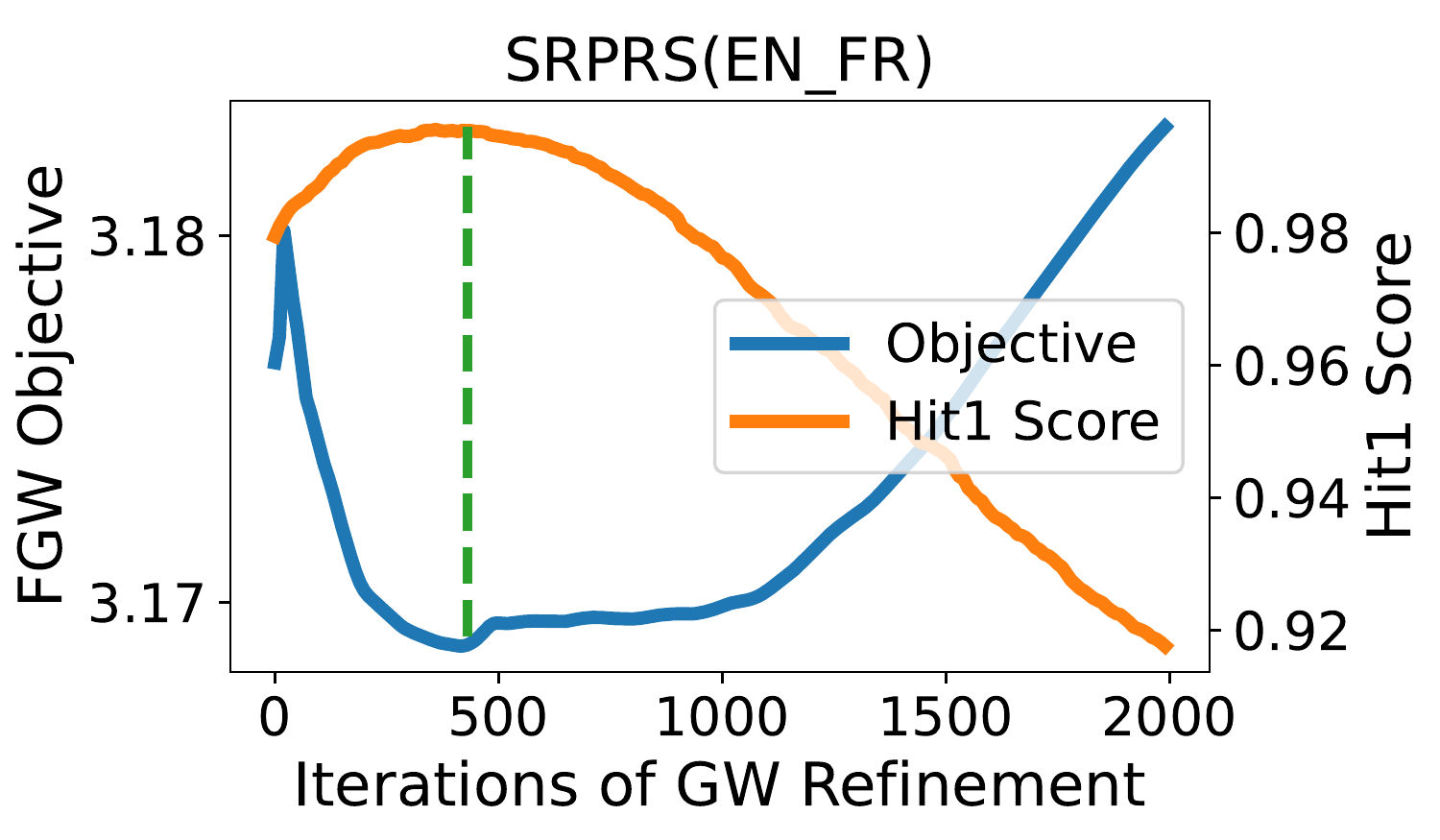} 
\caption{Visualization of the relationship between the objective function and alignment performance on SRPRS$_\textnormal{EN\_FR}$.}
\label{fig:exp3}
\end{figure}

\section{More Examples of the FGW Objective}\label{appendix:case}
Same as Section \ref{sec45}, in Figures \ref{fig:exp2} and \ref{fig:exp3}, we visualize the objective function in \eqref{eq:finalFGW} and the corresponding Hit1 score in GW refinement on DBP15K$_\textnormal{JA\_EN}$ without translation and attributes and SRPRS$_\textnormal{EN\_FR}$. The observation is consistent with Section \ref{sec45}. Two curves are highly correlated and the iteration corresponding to the minimum FGW objective value is approximately that to the maximum Hit1 score.

\section{Additional Results}
Several studies have pointed out that many entities in DBP15K can be directly matched by strings to obtain aligned entities \citep{attrGNN}. In light of this, we perform additional experiments on a hard test set split of DBP15K, as introduced in \citep{attrGNN}, to minimize the influence of name bias. Furthermore, to demonstrate that FGWEA's exceptional performance cannot be solely credited to the powerful LaBSE encoder, we use the mean pooling of \texttt{bert-base-multilingual-cased}\footnote{\url{https://huggingface.co/bert-base-multilingual-cased}} as FGWEA's new semantic encoder. The embedding matching accuracy for this encoder is only 16.1\% on the hard setting of DBP15K$_{\textnormal{ZH\_EN}}$. Nonetheless, FGWEA continues to achieve competitive results as shown in Table \ref{tab:app}, surpassing AttrGNN, the current top-performing method for this setting \citep{attrGNN}.

\begin{table}[t]
\centering
\small
\resizebox{\linewidth}{!}{
\begin{tabular}{ll|ccc}\toprule
Model & Dataset        & Hit1            & Hit10          & MRR               \\\midrule
&DBP15K$_\textnormal{ZH\_EN}$      &\textbf{0.756}	&\textbf{0.868}	&\textbf{0.796 }   \\
FGWEA&DBP15K$_\textnormal{JA\_EN}$      &\textbf{0.788}	&0.897	&\textbf{0.828}           \\
&DBP15K$_\textnormal{FR\_EN}$      &\textbf{0.983}	&\textbf{0.997}	&\textbf{0.988}           \\\midrule
&DBP15K$_\textnormal{ZH\_EN}$      &0.662 &0.818 &0.719    \\
AttrGNN&DBP15K$_\textnormal{JA\_EN}$      &0.774	&\textbf{0.903}	&0.821           \\
&DBP15K$_\textnormal{FR\_EN}$      &0.886	&0.956	&0.912           \\
\bottomrule
\end{tabular}
}
\caption{Results comparison bewteen FGWEA and AttrGNN on a hard setting of DBP15K.}\label{tab:app}

\end{table}

\end{document}